\pdfoutput=1

\documentclass[11pt]{article}

\usepackage[preprint]{coling}

\usepackage{times}
\usepackage{latexsym}
\usepackage[T1]{fontenc}

\usepackage[utf8]{inputenc}
\usepackage{microtype}

\usepackage{inconsolata}
\usepackage{microtype}
\usepackage{CJKutf8}
\usepackage{amsfonts, amsmath, amssymb}
\usepackage[english]{babel}
\usepackage{calc}
\usepackage{ifthen}
\usepackage{tikz}
\usepackage{url} 
\usepackage{tabularx}
\usepackage{float}
\usepackage{rotating,booktabs,multirow}
\usepackage{adjustbox}
\newcommand\ignore[1]{}

\usepackage{color}
\usepackage{booktabs, multirow} 
\usepackage{soul}
\usepackage{array}
\usepackage{color, colortbl}
\usepackage{arydshln}

\usepackage{listings}

\title{DualCoTs: Dual Chain-of-Thoughts Prompting for Sentiment Lexicon Expansion of Idioms}

\author{Fuqiang Niu$^{1}$\thanks{Equal contribution.}, Minghuan Tan$^{2}$\footnotemark[1], Bowen Zhang$^{1}$\thanks{Corresponding author.}, Min Yang$^{2}$\footnotemark[2] \and Ruifeng Xu$^{3}$ \\
        $^{1}$The College of Big Data and Internet, Shenzhen Technology University, China \\ 
        $^{2}$Shenzhen Institute of Advanced Technology, Chinese Academy of Sciences, China\\
        $^{3}$Harbin Institute of Technology, China \\
        nfq729@gmail.com \ \ \ \ \{mh.tan,min.yang\}@siat.ac.cn \\
        zhang\_bo\_wen@foxmail.com \ \ \ \  xuruifeng@hit.edu.cn}

\begin{document}
\maketitle

\begin{CJK*}{UTF8}{gbsn}

\maketitle
\begin{abstract}

Idioms represent a ubiquitous vehicle for conveying sentiments in the realm of everyday discourse, rendering the nuanced analysis of idiom sentiment crucial for a comprehensive understanding of emotional expression within real-world texts. Nevertheless, the existing corpora dedicated to idiom sentiment analysis considerably limit research in text sentiment analysis. In this paper, we propose an innovative approach to automatically expand the sentiment lexicon for idioms, leveraging the capabilities of large language models through the application of Chain-of-Thought prompting. To demonstrate the effectiveness of this approach, we integrate multiple existing resources and construct an emotional idiom lexicon expansion dataset (called \textbf{\textsc{EmoIdiomE}}), which encompasses a comprehensive repository of Chinese and English idioms. Then we designed the Dual Chain-of-Thoughts (\textbf{DualCoTs}) method, which combines insights from linguistics and psycholinguistics, to demonstrate the effectiveness of using large models to automatically expand the sentiment lexicon for idioms. Experiments show that DualCoTs is effective in idioms sentiment lexicon expansion in both Chinese and English. For reproducibility, we will release the data and code upon acceptance.

\end{abstract}


\section{Introduction}
\label{sec:intro}



Idioms are commonly used to imply a specific evaluation or emotional stance towards the things they represent~\cite{nunberg1994idioms}. 
Published sentiment lexicons suggest that idioms typically carry sentiment. In the case of English idioms, SLIDE~\cite{jochim-etal-2018-slide} analyzed a lexicon corpus of 5,000 frequently used idioms selected from Wiktionary, where over 40\% of the entries were labeled as sentiment-bearing. For Chengyu (成语), a type of frequently used Chinese idioms, out of the 40,000 commonly recognized items, 13,000 were annotated with at least one emotion in the Chinese Affective Lexicon Ontology (CALO)~\cite{xu2008constructing}.


Previous research has examined how idioms can impact sentiment analysis. 
In English, \citet{WILLIAMS20157375} studied the role of idioms in automated sentiment analysis approaches and demonstrated that including idioms as features improves traditional sentiment analysis results. \citet{spasic-etal-2020-taffc-idiom} have described an automated method to enhance sentiment analysis with idiom-based features. 
As for Chinese Chengyu, \citet{xie2014construction} proposed an unsupervised framework that utilizes Chinese idiom resources to develop a domain-independent sentiment classifier. This classifier was subsequently enhanced for domain-specific sentiment classification using labeled data.

However, the sentiment lexicons for idioms have limited entries and may become outdated due to infrequent updates. This means that many recognized idioms and newly coined idioms are not included, and their sentiments are not annotated. Consequently, it is important to investigate how we can automatically predict the sentiment labels of idioms in order to enhance existing idiom sentiment lexicons. 


With the advancement of large language models (LLMs) such as Bloom~\cite{workshop2023bloom}, GPT4~\cite{openai2023gpt4}, ChatGLM~\cite{zeng2023glmb}, and Llama~\cite{touvron2023llama}, their capabilities in reasoning, planning, and interaction are becoming increasingly strong. 
An example of this is seen in ChatGPT~\cite{gilardi2023chatgpt}, which outperforms human crowd-workers in various annotation tasks such as relevance, stance, topics, and frames detection.
In large language models, the ability to reason naturally emerges through a method called Chain-of-Thought (CoT). 
CoT prompts LLMs to generate a coherent series of intermediate reasoning steps leading to the final answer to the original question. 
CoT-based techniques have been successfully applied to sentiment analysis.\citet{Fei2023ReasoningIS} introduces the Three-hop Reasoning (THOR) CoT framework, which mimics the human-like reasoning process for implicit sentiment analysis (ISA). 
THOR achieves impressive performance improvements in both supervised and zero-shot setups.
Considering the significance of sentiment lexicon expansion for idioms in the fields of sentiment analysis and opinion mining, we propose using existing resources and more refined CoTs to effectively elicit LLMs for expanding sentiment lexicons for idioms.





In the fields of linguistics and psycholinguistics, the dual idiom representation model suggests that idiomatic expressions can be viewed as both ``long words'' and compositional phrases~\cite{TITONE19991655}. 
This model has prompted the exploration of two primary approaches for representing and processing idioms: the compositional approach~\cite{chomsky1980rules,fraser1970idioms,van1992incremental,bobrow1973catching,gibbs1980spilling,swinney1979access}) and the noncompositional approach. 
The compositional approach places its emphasis on the individual constituents comprising an idiom, dissecting it into its constituent words, and treating it as a sequence of interconnected lexical elements. In contrast, the noncompositional approach perceives idiomatic expressions as holistic entities, wherein the meaning emerges from the idiom as an integrated whole, transcending the sum of its individual parts. 

Building upon the dual idiom representation model, we introduce the Dual Chain-of-Thoughts (\textbf{DualCoTs}) prompting method. 
DualCoTs consist of two distinct chains: the \textit{literal} chain and the \textit{etymological} chain. 
The literal chain aims to simulate a literal understanding of idiomatic expressions by treating them as compositional phrases. It focuses on the surface forms and analyzes the constituent words of the idiom.
Conversely, the ``etymological'' chain explicitly incorporates directives to delve into the origin and historical context of idiomatic expressions.
By doing so, LLMs are encouraged to perceive and interpret idiomatic expressions as integrated units, thereby capturing the essence of their idiomatic nature.
These two chains, the literal and etymological, are combined to predict the sentiment associated with a particular idiom. By incorporating both approaches, the DualCoTs method provides a comprehensive and nuanced analysis of idiomatic expressions.


To facilitate the verification of our proposed methods and benchmark sentiment lexicon expansion for idioms, we construct a emotional idiom lexicon expansion dataset (\textbf{\textsc{EmoIdiomE}}) consisting of both English language and Chinese language.
\textsc{EmoIdiomE} pairs each idiom with multiple sentences to indicate its usage.
The idioms are divided into \textit{Train}, \textit{Dev} and \textit{Test} splits.
Therefore, the dataset can be used for supervised learning.
There is also an extra \textit{Unlabelled} split for human evaluation.
The dataset can not only used for sentiment lexicon expansion but also emotional lexicon expansion.
But in this paper, we will focus on sentiment lexicon expansion.

We conduct experiments over \textsc{EmoIdiomE} using both prompt-learning and CoT-based prompting methods.
Our CoT-based prompting methods take advantage of idioms' properties~(fixedness, origins) and integrate instructions accordingly to address these different aspects.
Through our experiments, we discover that idioms' properties have a discernible impact on the performance of the CoTs methodology.
In order to comprehensively address this issue, we add one control set to make deeper analysis how each property of idioms may influence the results.
Furthermore, We also apply our method to unlabelled idioms for human annotation. The annotated results of this annotation process were characterized by a notably high level of accuracy. This underscores the efficacy of our approach in the context of sentiment lexicon expansion for Idioms. 
In summary, our work makes the following contributions:
(1) We construct the \textsc{EmoIdiomE} dataset, which can be used not only for sentiment lexicon expansion of idioms but also finds application in the broader domain of sentiment analysis tasks.
The dataset covers two languages, Chinese and English, thus rendering it applicable for a wide range of cross-lingual research endeavors.
(2) We proposed the DualCoTs prompting method, specifically designed for idioms, to expand idiom sentiment lexicons and address the existing limitations.
(3) We demonstrate DualCoTs' effectiveness in sentiment judgment through experiments, and show its applicability to expanding sentiment lexicons for unlabeled idioms.

\section{Related Work}
\label{sec:related}

\subsection{Automatic Construction of Sentiment Lexicons}

In recent years, the construction of sentiment lexicons has evolved with diverse approaches tailored to various domains and contexts. For instance, \citet{lu-etal-www-2011-automatic} developed an optimization framework that integrates different information sources to learn a sentiment lexicon that adapts to domain specifics and contextual aspects of unlabeled opinion texts. Additionally, \citet{hu-etal-2013-www-unsupervised} explored unsupervised sentiment analysis by leveraging emotional signals such as emoticons and product ratings, which underscore the feasibility of extracting sentiment data without direct supervision.

Further developments in deep learning have refined the methods used for constructing sentiment lexicons. Notably, phrase-level sentiment classification techniques~\cite{tang-etal-2014-building} and neural architectures that integrate sentiment supervision at both the document and word levels~\cite{wang-xia-2017-sentiment} have markedly improved the quality and utility of sentiment-aware embeddings. In addition, domain-specific methodologies, such as the emoji sentiment lexicon developed by \citet{kimura-etal-2017-asonam-automatic} and the topic-adaptive sentiment lexicon by \citet{deng-etal-2019-taslp-sentiment}, cater to the variability of sentiment expressions across different topics and domains, thus providing more sophisticated tools for specialized sentiment analysis applications. Furthermore, \citet{prochnow-etal-2024-idem} have introduced the idioms with emotions dataset, employing pre-training models in supervised learning settings to categorize idioms effectively.

\subsection{Chain-of-Thought Prompting}

A Chain-of-Thought~\cite{wei2023chainofthought} prompts sufficiently LLMs to generate a coherent series of intermediate reasoning steps that lead to the final answer to the original question, thereby naturally eliciting the reasoning abilities of these models.

Chain-of-Thought prompting methods have been applied across various fields in natural language processing, and new techniques have been proposed to tackle specific challenges that prevent LLMs from operating rationally. To enhance the reasoning ability of LLMs, \citet{wang2023selfconsistency} first samples a diverse set of reasoning paths and then selects the most consistent answer by marginalizing out all possible reasoning paths. In sentiment analysis, the Three-hop Reasoning (THOR)~\cite{Fei2023ReasoningIS} CoT framework simulates the human-like reasoning process for implicit sentiment analysis (ISA). Moreover, \citet{duan2024implicit} leverage the CoT approach to analyze the implicit aspects and opinions in texts, thereby facilitating a more nuanced analysis of sentiments.


\section{Dataset Construction}
\label{sec:data}

\subsection{Existing Datasets}

The construction of a lexicon expansion dataset requires the acquisition of appropriate training data that includes both the contextual usage of idioms and corresponding affective lexicons. To meet this requirement, we have pursued distinct strategies for Chinese and English idioms: 
For Chinese idioms, we utilize the \textsc{\textbf{ChID}}\cite{zheng-etal-2019-chid}, which provides contexts from reading comprehension, along with the \textbf{CALO}\cite{xu2008constructing} for emotion and sentiment annotations. For English idioms, we combine resources from \textsc{\textbf{Idioment}}\cite{WILLIAMS20157375}, \textsc{\textbf{IdiomParaphrases}}\cite{pershina-etal-2015-idiom}, and \textsc{\textbf{SLIDE}}\cite{jochim-etal-2018-slide} for sentiment annotations and retrieve contexts from large corpora such as the British National Corpus(BNC)~\cite{BNC} and the One Billion Word Corpus.

\paragraph{CALO} The Chinese Affective Lexicon Ontology (CALO) was developed to support Affective Computing (AC) in the Chinese language. The creation of CALO was influenced by mainstream emotional classification research~\cite{ekman1992argument}, combined with traditional Chinese emotional categories. However, the category `enjoyment' (乐) did not adequately describe some positive emotions such as `respect' (尊敬) and `belief' (相信). Therefore, an additional category, `good'(好), was introduced. In total, CALO contains seven main categories, each with several subcategories graded by intensity levels ${1,3,5,7,9}$. In this paper, we refer to the seven main categories as coarse-grained emotions and the twenty-one subcategories as fine-grained emotions. The mapping between coarse-grained and fine-grained emotions is detailed in Appendix~\ref{appendix1}.

\begin{table*}[t]
\centering
\begin{adjustbox}{max width=0.7\linewidth}
\begin{tabular}{lrrrrrrrrrrr}
\toprule
& &\multicolumn{4}{c}{ZH} & &\multicolumn{4}{c}{EN} \\
\cmidrule{3-6}
\cmidrule{8-11}
& &Train &Dev &Test &Unlabelled & &Train &Dev &Test &Unlabelled \\\midrule
Idioms & &1,708 &571 &575 &1,006&  &2,981& 1,000& 993 & 1,309 \\\midrule
K=1 & &1,708 &571 &575 &1,006&  &2,981& 1,000& 993 & 1,309 \\
K=4 & &6,826 & 2,280& 2,300&4,024 & &11,898 & 3,992& 3,966&4,512 \\
K=8 & & 13,643& 4,554& 4,594& 8,048& & 23,761&7,969 & 7,907&7,933 \\
K=16 & & 27,242& 9,089& 9,160& 16,087& &47,210 &15,838 & 15,698&13,313 \\
All & &295,406 &95,283 &99,003 & 159,384& &47,210 &15,838 & 15,698&13,313 \\
\bottomrule
\end{tabular}

\end{adjustbox}
\caption{\label{tab:emoidiome} Statistics of the \textsc{EmoIdiomE} dataset.}
\end{table*}

\paragraph{\textsc{ChID}} \textsc{ChID} is a large-scale Chinese Idiom Dataset with the goal to facilitate the study of Chengyu comprehension using deep learning models.
The dataset was created in the ``cloze'' style.
The text includes novels and essays from the Internet and news articles.
To construct the candidate answer set for each masked Chengyu, the authors considered synonyms, near-synonyms and other Chengyu either irrelevant or opposite in meaning to the ground truth Chengyu.

\paragraph{\textsc{Idioment}} \citet{WILLIAMS20157375} collected a set of 580 idioms that are relevant to sentiment analysis, i.e. the ones that can be mapped to an emotion.
A total of 2900 annotations were collected for all 580 idioms with 5 annotations per idiom. 
A total of 8610 annotations were collected for all sentences in the corpus with at least 3 annotations per sentence. 

\paragraph{\textsc{IdiomParaphrases}} \citet{pershina-etal-2015-idiom} annotated 1.4K idiom paraphrase pairs for the task short-text paraphrase identification with idiomatic expressions. 
The idioms are collected from a site\footnote{\url{http://www.usingenglish.com}} for English learners.
Each idiom has a unique description giving a clear explanation of the idiom’s meaning.
Although there's no sentiment labels on this dataset, we use the dataset to expand our vocabulary of idioms and treat them as unlabelled data.

\paragraph{SLIDE} Sentiment Lexicon of IDiomatic Expressions is an English corpus on sentiment annotation of idiomatic multiword expressions collected using crowdsourcing. 
It collects 10 annotations for each idiom and the aggregated label is shown to have good agreement with expert annotations. 
SLIDE is much larger than previous idiom lexicons which includes 5,000 frequently occurring idioms, as estimated from a large English corpus. 
The idioms were selected from Wiktionary, and over 40\% of them were labeled as sentiment-bearing.

\subsection{\textsc{EmoIdiomE} Dataset}

To validate the effectiveness of our proposed method in the automated construction of sentiment lexicons, we construct a dataset, named \textsc{EmoIdiomE}, encompassing idioms in both the Chinese and English languages. 
During the dataset construction process, we notice that the number of passages for each idiom is largely skewed which ranges from several to hundreds. 
This disparity in passage counts could potentially lead to overfitting to more commonly used idioms.
To mitigate this issue, we have taken measures to construct a more balanced dataset, where each idiom is coupled with at most $K$ passages. 
If the number of passages for an idiom is less than $K$, all the passages are used.
In this work, we choose $K=1,4,8,16$.
For both English language~(EN) and Chinece language~(ZH) in this dataset, the three splits \emph{Train}, \emph{Dev} and \emph{Test} contain only idioms that have labels from existing lexicons.
We also include all the \emph{Unlabelled} idioms in this dataset for further usage.

\paragraph{ZH} To construct an affection-oriented dataset for Chinese Idioms, we reuse the training split of ChID which contains 520k passages covering 3848 distinct idioms.
Among the 3848 idioms, there are 2864 idioms which have been labeled in CALO.
There is only one fine-grained emotion type `NK~(envy, 妒忌)' is not covered in ChID.
Therefore, we further add 12 `NK' idioms with 156 passages\footnote{We manually crawled them from literature works.} to the dataset.

\paragraph{EN} For English idioms, we combine \textsc{Idioment}, \textsc{IdiomParaphrases} and \textsc{SLIDE}.
However, these datasets does not provide us sentences that contain the idioms.
As a result, we extract sentences from two corpora BNC and One Billion Word~\cite{chelba2013one} for each idiom.
Similar to the practices used by SLIDE, we preprocess idioms that have personal pronouns or possessive adjectives when retrieving sentences.

Table \ref{tab:emoidiome} provides a comprehensive overview of our \textsc{EmoIdiomE} dataset. This dataset encompasses a total of 7,828 idioms. Specifically, there are 2,854 labeled data entries for Chinese idioms, along with 1,006 unlabeled entries. In the case of English idioms, there are 4,974 labeled data entries and 1,309 unlabeled entries. Impressively, the dataset comprises a substantial number of example sentences, totaling 741,135 instances. The extensive coverage of both the number of idioms and example sentences in this dataset underscores its versatility and potential for broad applicability in the field related to idioms. This breadth allows the dataset to address a wide spectrum of issues and challenges associated with idiom.

\section{Dual Chain-of-Thoughts Prompting}

In this section, we begin by introducing several baseline methods derived from prompt learning and CoT-based prompting. Furthermore, we present a pre-trained model approach, which serves as a point of comparison with the CoT methodology. Following this, we proceed to introduce our instructed prompting approach for the DualCoTs method.

\begin{figure*}[t]
\centering
\includegraphics[width=0.8\linewidth]{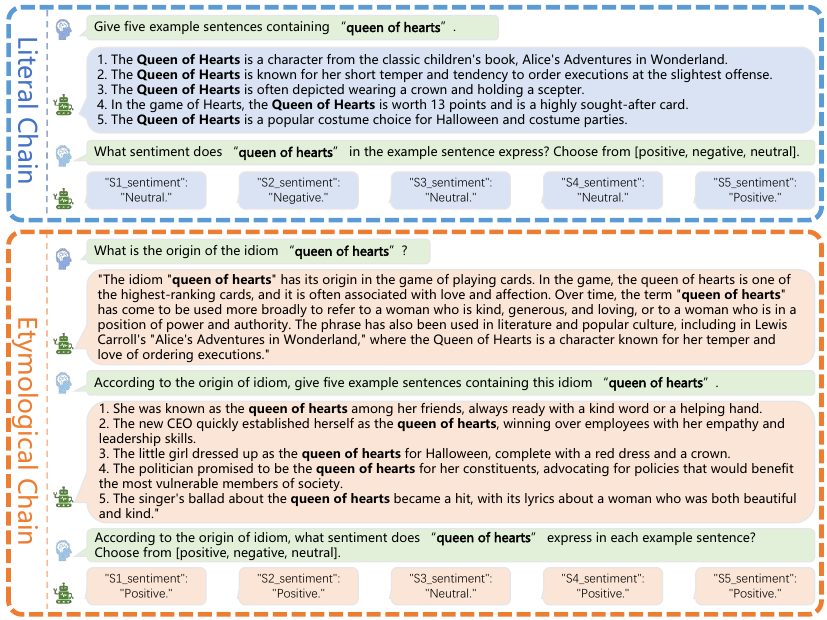}
\caption{Dual Chain-of-Thoughts. In this example, if we only rely on the \textit{literal} chain, the prompted example sentences may not focusing the idiomatic aspect of the phrase. The introduction of the \textit{etymological} chain can alleviate the issue by focusing more on the idiomatic usage. The predicted sentiment will shift from \textit{neutral} to \textit{positive}.} 
\label{fig:dualcots}
\end{figure*}

\subsection{Baseline Methods}

\paragraph{Direct Inquiry} This method involves directly soliciting sentiment predictions from LLMs based on the idiomatic expressions they have learned, without context beyond the idiom's surface form.

\paragraph{Idiom Inquiry} This approach enhances sentiment prediction by informing LLMs that a phrase is an idiom, prompting them to retrieve relevant background knowledge and insights.

\paragraph{Usage Inquiry} This method solicits sentiment assessment of an idiom within a contextually provided example sentence, facilitating a better understanding of the idiom's commonly used meaning.

\paragraph{Origin Inquiry} Understanding an idiom's origins, often rooted in historical or cultural contexts, is crucial, and this inquiry method focuses on enhancing LLMs' comprehension of idioms for improved sentiment predictions.

\paragraph{Origin and Usage Inquiry} This comprehensive method combines inquiries into an idiom's origin and usage, sequentially prompting LLMs to provide a deeper understanding of the idiom's sentiment in context.

\paragraph{PromptT5} To contrast with the CoT method, we have compared it to a well-performing prompt learning~\cite{10.1145/3560815} method. Prompt learning directly adapts pre-trained language models~(PLMs) to downstream tasks using either generation-based or mask-filling-based approaches, and can achieve promising performances on various tasks.
We use the OpenPrompt~\cite{ding-etal-2022-openprompt} toolkit to implement the prompt model for sentiment lexicon expansion for idioms.
Our pipeline consists of a template and a verbalizer.
We use a template is ``<pre text><idiom><post text>. <idiom> is <mask>'', where
the token <pre text> stands for the original text before the idiom and <post text> stands for the original text after the idiom.
Our verbalizer uses the label words \textit{positive}, \textit{negative} and \textit{neutral}.
For English, we use the T5~\cite{2020t5} model\footnote{\url{https://huggingface.co/t5-base}}.
For Chinese, we use a T5 model\footnote{\url{https://huggingface.co/uer/t5-v1_1-base-chinese-cluecorpussmall}} released by \citet{zhao-etal-2019-uer}.

\subsection{DualCoTs}
Our Dual Chain-of-Thoughts~(DualCoTs) serves to emulate the dual representation model of idioms using large language models. Within the DualCoTs framework, two distinctive chains operate in tandem:

 (i) The \textit{literal} chain is to directly elicit judgement of sentiment from LLMs with surface forms of the idioms.
 (ii) The \textit{etymological} chain employs idiom-aware instructions before prompting LLMs for their sentiment assessments. This added layer of instruction aims to enhance the LLMs' understanding of the idiomatic context and its potential impact on sentiment analysis.

To investigate if LLMs rely on usage of idioms when making the judgements, we further extend each chain by injecting instructions of giving example sentences containing the idiom.

Specifically, as illustrated in Figure~\ref{fig:dualcots}, there are two separate chains with different instructions.
Within the \textit{literal} chain, we have divided the process into two steps. 
(1) We initiate the process by employing a usage inquiry approach to prompt LLMs to generate multiple example sentences that effectively convey the sentimental implications of the idiom. This initial step aims to elucidate the extent of knowledge retained by LLMs regarding the idiom.
(2) Subsequently, leveraging the example sentences provided in the first step, we proceed to extract the sentimental connotations conveyed by the idiom within each sentence. This two-step methodology enables a more comprehensive understanding of the idiomatic expressions and their associated sentiments across various contextual applications.

Within the \textit{etymological} chain, we have divided the process into three steps. 
(1) Commencing with an inquiry about the origins of the idiom, we explicitly engage with LLMs to prompt them for insights into the idiom's historical or cultural roots. This initial step serves to acquaint LLMs with the idiomatic nature of the provided phrase and its original usage as an idiom.
(2) Building upon the knowledge of the idiom's origins, we prompt LLMs to provide practical examples that stem from these origins. This step is instrumental in enhancing LLMs' understanding of the idiom's usage and enhances their cognitive capabilities in the context of idiomatic expressions.
(3) Drawing from the knowledge of the idiom's origins and the provided examples, we inquire about the emotional meanings conveyed by the idiom, thus capturing its emotional nuances within an idiomatic context. This three-step process collectively aims to deepen LLMs' comprehension of the idiom's unique characteristics and its emotional dimensions.

Each chain generates five sentiment predictions. The ultimate sentiment prediction is determined through a voting mechanism that combines the results obtained from both chains.

\begin{table*}[!htp]\centering
\resizebox{\linewidth}{!}{
\begin{tabular}{lccccccccccccccccccc}\toprule
& &\multicolumn{2}{c}{PromptT5} & &\multicolumn{4}{c}{ChatGLM-6B} & & \multicolumn{3}{c}{LLama2-70b}& &\multicolumn{4}{c}{ChatGPT} & \\
\cmidrule{3-4}\cmidrule{6-9}\cmidrule{11-13}\cmidrule{15-18}
& &\multicolumn{2}{c}{EmoIdiomE} & &\multicolumn{2}{c}{EmoIdiomE} & &Idioment    &&EmoIdiomE &&Idioment& &\multicolumn{2}{c}{EmoIdiomE} & &Idioment & &Avg.\\
\cmidrule{3-4}\cmidrule{6-7}\cmidrule{9-9}\cmidrule{11-11}\cmidrule{13-13}\cmidrule{15-16}\cmidrule{18-18}
& &ZH &EN & &ZH &EN & &EN    && EN&&EN& &ZH &EN & &EN  &\\\midrule
K=1 & &73.2 &63.8 & &- &- & &-    && -&&-& &- &- & &-  &&-\\
K=4 & &73.7 &64.8 & &- &- & &-    && -&&-& &- &- & &-  &&-\\\midrule
Direct Inquiry & &- &- & & 35.5&38.9 & &38.6    && 60.5&&75.8& &46.6 &\underline{66.0} & &70.3  &&54.0\\
Usage Inquiry & &- &- & & 30.6 & 38.9 & &\underline{48.1}    && 55.7&&78.4& & 68.9&59.1 & &\underline{78.8}  &&57.3\\
Idiom Inquiry & &- &- & & 35.3&26.1 & &28.4    && \underline{63.3}&&77.0& &45.7 &\textbf{71.1} & &74.3  &&52.7\\
Origin Inquiry & &- &- & & 37.7&\underline{46.1} & &43.6    && 50.4&&\underline{79.0}& &69.7 &43.5 & &\underline{78.8}  &&56.1\\
Origin and Usage & &- &- & & \underline{39.3}& 41.7 & &45.7    && 58.6&&73.2& &\underline{69.9} &58.4 & &78.6  &&\underline{58.2}\\
\midrule
DualCoTs & &- &- & &\textbf{40.7} &\textbf{46.6} & &\textbf{49.7}    && \textbf{68.2}&&\textbf{80.8}& &\textbf{71.8} &60.1 & &\textbf{80.0}  &&\textbf{62.2}\\
\bottomrule
\end{tabular}
}
\caption{\label{tab:result} Experiment results on the test split of the \textsc{EmoIdiomE} dataset and \textsc{Idioment}. The performance metrics are reported in terms of accuracy, with Avg. representing the average score across all models.}
\end{table*}

\section{Experiments}
\label{sec:exp}
In this section, we perform comprehensive experiments on our \textsc{EmoIdiomE} dataset. In addition to the \textsc{EmoIdiomE} dataset, we tested several methods on the \textsc{Idioment} dataset to assess our method generalizability. Furthermore, we present the results of our method when applied to unlabelled data, thus substantiating the efficacy of our approach in the automatic expansion of idiom sentiment lexicons.

We  conduct experiments with ChatGPT (gpt-3.5-turbo\footnote{\url{https://platform.openai.com/docs/models/gpt-3-5}}), LLaMA (LLama 2-70b\footnote{\url{https://huggingface.co/meta-llama/Llama-2-70b-chat-hf}}) and ChatGLM(ChatGLM-6b\footnote{\url{https://huggingface.co/THUDM/chatglm-6b}}), which are popular and powerful LLMs. Additionally, we opted for ChatGLM-6B, primarily due to its excellent performance in the context of the Chinese language.

\subsection{Sentiment Lexicon Expansion for Idioms} 
\paragraph{Intrinsic Evaluation on Labelled Data} 

We evaluate how each method performs over \textit{Test} split of \textsc{EmoIdiomE} and list all experiment results in Table~\ref{tab:result}.
We first include results for PromptT5 using one or four sentences per idiom.
Then we list all the CoT-based methods, Direct Inquiry, Usage Inquiry, Idiom Inquiry, Origin Inquiry and Origin \& Usage Inquiry.
Our DualCoTs is shown at the bottom.

Generally speaking, we have the following main findings:

(1) PromptT5 with supervision is still competitive compared to prompting methods over LLMs. But the gap is very small.
   Using more example sentences for prompt learning can help improving the accuracy of sentiment expansion.
   
(2) The use of example sentences and the introduction of idiom origin in the prompt method show significant improvements, which also serve as evidence of the effectiveness of our DualCoTs.

(3) The size of large language models is one significant factor to the prompting performance.
   ChatGPT and LLaMA works consistently better than ChatGLM-6B. The reason for LLaMA potentially outperforming ChatGPT in the results might be due to the fact that LLaMA was fine-tuned using only English data, which could lead to better consistency and, consequently, better results compared to ChatGPT.

It's worth noting that the difference between Chinese idioms and English Idioms on \textsc{EmoIdiomE}.
Chinese idioms are all Chengyu, a commonly used idiom type, which have higher fixedness and can be easily recognized.
English idioms tends to be more flexible~\cite{fazly-stevenson-2006-automatically} that have multiple senses and may only be potential idiomatic expressions~\cite{tayyar-madabushi-etal-2021-astitchinlanguagemodels-dataset}. 

We also notice that, for ChatGPT, the Idiom Inquiry surpasses the Origin Inquiry by a large margin. 
We suspect this is due to the test set of \textsc{EmoIdiomE} contains many idioms with no clear origins.
For example, the test set of \textsc{EmoIdiomE} contains frequently used terms like ``\textit{come down with}'' and ``\textit{come in}'' which have no clear origins and ChatGPT cannot offer related information through Origin Inquiry for the judgement of their sentiment.
To further verify this, we use \textsc{Idioment} as a control set in which idioms have clearer origins compared with \textsc{EmoIdiomE}.

Combining results from ZH of test set on \textsc{EmoIdiomE} and \textsc{Idioment}, we can conclude that for idioms that have strong idiomatical background and clearer origins, DualCoTs can achieve consistent improvement over other CoT-based prompting methods.

\paragraph{Evaluation on Unlabelled Data} 

We randomly select 50 idioms for each language from the \textit{Unlabelled} split of \textsc{EmoIdiomE}.
Then we use the best method DualCoTs over \textit{gpt-3.5-turbo} to predict their sentiments.
Two annotators are asked to evaluate the predictions.
The prediction accuracy and annotation consistency are listed in Table~\ref{tab:unlabelled}.

\begin{table}[!htp]\centering
\begin{adjustbox}{max width=0.5\linewidth}
\begin{tabular}{lccc}\toprule
&EN &ZH \\
\midrule
Accuracy &86.0 & 84.0 \\
Consistency & 86.0 & 82.0 \\
\bottomrule
\end{tabular}
\end{adjustbox}
\caption{Prediction accuracy and annotation consistency (percentage agreement) for a subset of \textit{unlabeled} data.} \label{tab:unlabelled}
\end{table}

The annotation results indicate that DualCoTs can be used for automatically expanding sentiment lexicons for idioms with high reliability.
The method has potential usage in lexicon construction to reduce annotation costs.

\paragraph{Human Evaluation on New Idioms} 

To test the performance of DualCoTs over out-of-domain phrase sentiment prediction, we use most recent cyberwords
that do not exist in our vocabulary and manually assess their sentiment.
These newly-coined idioms emerge within a short period online that LLMs have smaller possibilities to have seen their annotations or even usages.
We give two examples below.

For Chinese language, we utilize the idiom ``心满离''~(heart, full, leave) which is a recent buzzword online.
Although LLMs cannot determine the exact origin of this idiom, they are able to develop a certain understanding of it. 
Here is an example sentence for the idiom, ``这部电影的结尾令人感动，主人公与自己的过去告别，但他的心中却充满了一种心满离的满足感'' (The ending of this movie is moving, as the protagonist bids farewell to their past, yet their heart is filled with a sense of ``心满离'' satisfaction), which conveys positive sentiment that aligns with the sentiment associated with internet buzzwords.

For English language, we utilize the idiom \textit{Composter Syndrome}\footnote{\url{https://www.urbandictionary.com/author.php?author=peepeecomposter\%20}}.
Currently, \textit{Composter Syndrome} refers to a psychological phenomenon experienced by individuals who have an intense desire to efficiently organize and manage composting processes.
In our DualCoTs, when exploring \textit{Composter Syndrome} through the \textit{idiomatic} Chain, LLMs cannot provide information related to idioms.
However, in the \textit{literal} Chain, LLMs tend to associate \textit{Composter Syndrome} with positive impressions and provide an example sentence: ``Her `composter syndrome' became evident when she proudly showed off her meticulously organized composting system, complete with color-coded bins and a spreadsheet tracking decomposition rates.''
This example sentence conveys a positive sentiment just as the idiom \textit{Composter Syndrome} in internet discourse.

\subsection{Discussion over Generated Origins}
\label{sec:origin}

\begin{figure}[t]
\centering
\includegraphics[width=0.9\linewidth]{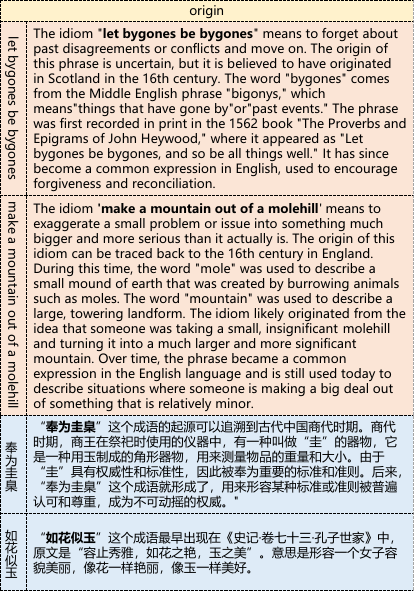}
\caption{Generated origins of idioms from LLMs. The English translations of the Chinese idioms can be found in Appendix~\ref{appendix2}.} 
\label{fig:origin}
\end{figure}

Large language models suffer a lot from factual consistency issues~\cite{tam2022evaluating}.
As our method relies much on the generation ability of LLMs, it is inevitable for the method to use unfaithfully generated text for inference.
To control the risk caused by this, we investigate the most probable misinformation that might intervene the reasoning chains in our method.
As far as we are concerned, the generation of origins is more challenging for LLMs as these resources are scarce even online.

The generated origins, see Figure~\ref{fig:origin}, usually contain two major types of information, the background of the idiom and its idiomatic meaning.
We find that the background information has low factual consistency in most cases.
For example, the background information for ``如花似玉''~(as pretty as flower and jade) generated from ChatGPT is the book ``《史记·卷七十三·孔子世家》''~(Records of the Grand Historian, Volume 73: The House of Confucius).
But in fact, it should be ``《诗·魏风·汾沮洳》''~(Poetry: Airs of Wei – Fen Marshlands).
However, the idiomatic meaning of the idiom is accurate.
Considering the judgement of sentiment of an idiom is more related to its idiomatic meaning, our method is still robust against such misinformation.

\section{Conclusion}
In this paper, we explore the task of automatic expansion of sentiment lexicons for idioms using Chain-of-Thought prompting over large language models. 
Through the integration of diverse existing resources, we construct the \textsc{EmoIdiomE} dataset, which stands out as one of the most comprehensive datasets in terms of idiom coverage. With an extensive collection of idiomatic examples, this dataset holds the potential to address a wide array of issues within the realm of idioms.
Our innovative chain-of-thought methodology, DualCoTs, has been adequate assessed through practical experiments to validate its effectiveness in the task of automatically expanding sentiment lexicons for idioms.

\section*{Limitations}

Our DualCoTs have made significant advancements in the field of Sentiment Lexicon Expansion of Idioms. However, it is important to acknowledge the limitations we have encountered during its implementation.

Firstly, we have observed that the \textit{origin}s generated by LLMs may lack factual consistency, as discussed in Section~\ref{sec:origin}. 
This may lead to potential inaccuracies in the judgment of idioms. 
How to ensure reliability and accuracy of the generated information remains a challenge.

Secondly, we have found that LLMs face difficulties in providing origins for newly-coined idioms. This restricts DualCoTs from handling these types of expressions. 

Lastly, LLMs have shown biases when dealing with flexible idioms. In the absence of contextual cues, the generated contexts may lean towards a particular sentiment. 
How to address these biases and ensuring a more balanced sentiment analysis is a significant challenge for our DualCoTs.

\section*{Ethics Statement}
The datasets utilized in this study are sourced from openly accessible repositories, ensuring transparency and compliance with open data practices. The models employed are open-source and also include services from LLMs provided by OpenAI and Meta AI. In deploying these models, we adhere strictly to their terms and policies, ensuring that our research practices uphold the ethical standards set forth by these organizations.

\bibliography{custom}

\appendix

\section{CALO Classification}
\label{appendix1}
\begin{figure}[!hpt]
\centering
\includegraphics[width=\linewidth]{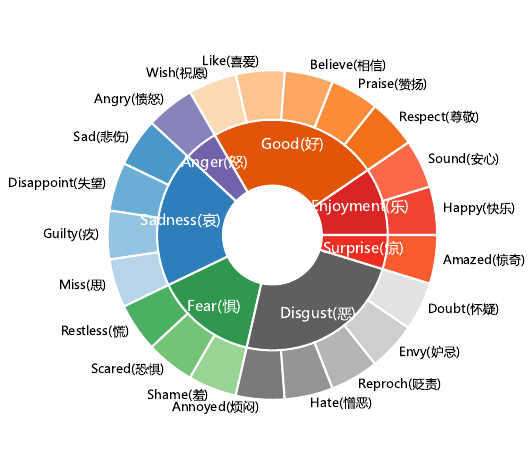}
\caption{Seven coarse-grained emotions in CALO, each coarse-grained emotion has different numbers of fine-grained emotions.} 
\label{fig:calo}
\end{figure}

\section{Case English Translation}
\label{appendix2}
\begin{center}
    \fcolorbox{black}{gray!10}{\parbox{0.95\linewidth}{奉为圭臬 (Hold up as an authoritative standard): The idiom ``奉为圭臬'' can be traced back to ancient China during the Shang dynasty. In the Shang period, there was an instrument used by the Shang kings in sacrificial ceremonies called a ``圭'', which was a jade-made, horn-shaped tool used to measure the weight and size of objects. Due to its authoritative and standardized nature, the ``圭'' was regarded as an important standard or criterion. Over time, the idiom ``奉为圭臬'' was formed, which is used to describe a certain standard or criterion that is universally recognized and respected, becoming an unshakable authority.}}
\end{center}

\begin{center}
    \fcolorbox{black}{gray!10}{\parbox{0.95\linewidth}{如花似玉 (As beautiful as a flower and as pure as jade): The idiom ``如花似玉'' first appeared in Records of the Grand Historian, Volume 73: The House of Confucius. The original text reads, "Her appearance is elegant and refined, as radiant as a flower, as beautiful as jade." It means to describe a woman's beauty, comparing her to the brilliance of flowers and the purity of jade.}}
\end{center}

\end{CJK*}{UTF8}{gbsn}
\end{document}